\documentclass{article}

\usepackage[nonatbib,final]{neurips_2023}

\usepackage{amsmath,amsfonts,bm}

\def\eqref#1{equation~\ref{#1}}

\def\1{\bm{1}}

\DeclareMathAlphabet{\mathsfit}{\encodingdefault}{\sfdefault}{m}{sl}
\SetMathAlphabet{\mathsfit}{bold}{\encodingdefault}{\sfdefault}{bx}{n}

\usepackage[utf8]{inputenc} 
\usepackage[T1]{fontenc}    
\usepackage{hyperref}       
\usepackage{url}            
\usepackage{booktabs}       
\usepackage{amsfonts}       
\usepackage{nicefrac}       
\usepackage{microtype}      
\usepackage{xcolor}         
\usepackage{multirow}
\usepackage[normalem]{ulem}
\usepackage{graphicx}
\usepackage{authblk}
\usepackage{makecell}
\usepackage{wrapfig}
\usepackage{tabularray}

\title{SEED-Data-Edit Technical Report: \\A Hybrid Dataset for Instructional Image Editing}

\begin{document}
\author{
\textbf{Yuying Ge$^{1\ast}$ \qquad Sijie Zhao$^{1\ast}$ \qquad Chen Li$^{2\ast}$ \qquad Yixiao Ge$^{1,2\dagger}$ \qquad Ying Shan$^{1,2}$\\} \vspace{-5pt} 

$^{1}$Tencent AI Lab \qquad $^{2}$ARC Lab, Tencent PCG\\
}

\maketitle

\begin{figure}[h!]
\vspace{-20pt}
\centering
\makebox[\textwidth][c]{\includegraphics[width=1.0\linewidth]{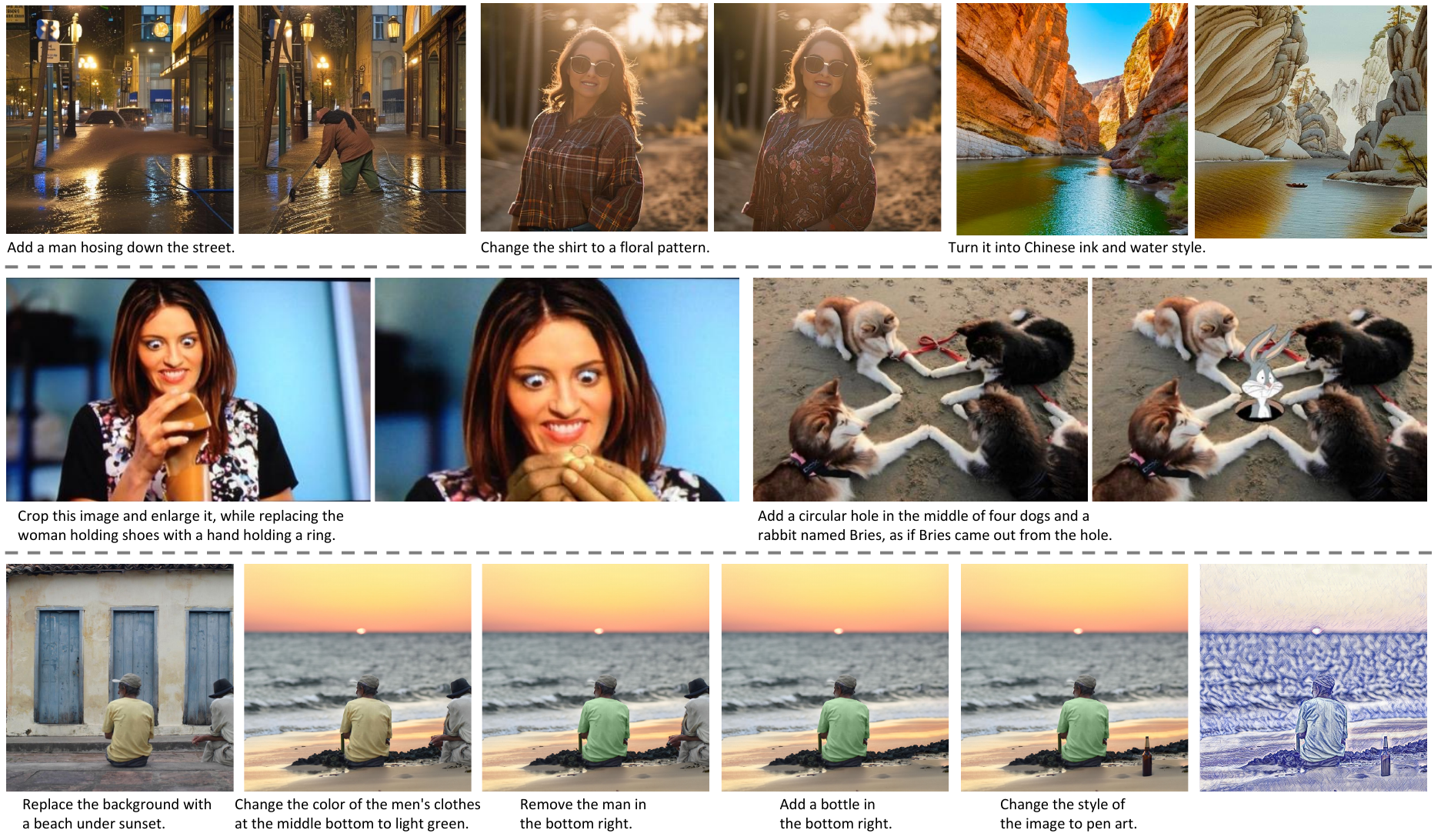}}%
\caption{Data examples of instruction-guided image editing in SEED-Data-Edit, which includes (1) High-quality editing data produced by an \textbf{automatic pipeline} (first row), (2) \textbf{Real-world scenario} data scraped from the internet that more accurately reflects user image editing intentions (second row), (3) High-precision \textbf{multi-turn} editing data annotated by Photoshop experts (third row).}
\label{fig:teaser_example}
\end{figure}

\begin{abstract}
In this technical report, we introduce SEED-Data-Edit: a unique hybrid dataset for instruction-guided image editing, which aims to facilitate image manipulation using open-form language. SEED-Data-Edit is composed of three distinct types of data: (1) High-quality editing data produced by an automated pipeline, ensuring a substantial volume of diverse image editing pairs. (2) Real-world scenario data collected from the internet, which captures the intricacies of user intentions for promoting the practical application of image editing in the real world. (3) High-precision multi-turn editing data annotated by humans, which involves multiple rounds of edits for simulating iterative editing processes.  The combination of these diverse data sources makes SEED-Data-Edit a comprehensive and versatile dataset for training language-guided image editing model. We fine-tune a pretrained Multimodal Large Language Model (MLLM) that unifies comprehension and generation with SEED-Data-Edit. The instruction tuned model demonstrates promising results, indicating the potential and effectiveness of SEED-Data-Edit in advancing the field of instructional image editing. The datasets are released in \url{https://huggingface.co/datasets/AILab-CVC/SEED-Data-Edit}.

\end{abstract}
 
 \renewcommand{\thefootnote}{\fnsymbol{footnote}}
 		\footnotetext[1]{Equal Contribution. } 
   \footnotetext[2]{Correspondence to \texttt{yixiaoge@tencent.com}.}

\section{Introduction}
Instruction-guided image editing~\cite{zhang2024magicbrush,brooks2023instructpix2pix,hui2024hq,sheynin2023emu,fu2023guiding} is an emerging field that empowers users to manipulate images using natural language instructions without complex descriptions or region-specific masks. This advancement significantly enhances the controllability and flexibility of image manipulation. However, compared with text-to-image generation~\cite{rombach2022high,betker2023improving,chen2023pixart}, instructional image editing is more challenging since it necessitates a comprehensive understanding of both language and visual content, and the capacity to handle iterative instructions while preserving both visual realism and semantic consistency. A significant hurdle in training models for instruction-guided image editing is the lack of high-quality, large-scale datasets, which are indispensable for learning a model to realize accurate interpretation and execution of editing instructions.

In this technical report, we introduce SEED-Data-Edit, which is a unique hybrid dataset meticulously crafted to tackle the challenges of instruction-guided image editing. As shown in Fig.~\ref{fig:teaser_example}, SEED-Data-Edit integrates three distinct types of data, providing a rich and versatile resource for training language-guided image editing models. The first part, automated pipeline-generated editing samples, ensures a substantial volume of diverse image editing pairs. The second part, real-world scenario data, encapsulates the complexities of user intentions, promoting the practical application of image editing in open-form context. The third part, human-annotated multi-turn editing data simulates iterative editing processes, enabling models to learn multiple rounds of image editing.

Specifically, for the first part of SEED-Data-Edit, we employ two automated pipelines to produce millions of editing pairs as shown in Fig.~\ref{fig:pipeline}. In pipeline (a), an object is initially segmented and subsequently removed using an inpainting module. This process results in a set of ``Remove'' and ``Add'' editing samples, where the ``Add'' samples are generated by 
reversing the ``Remove'' operation. In pipeline (b), we employ an image-guided text-to-image generation model to create source and target images based on the original image, the source caption and the target caption after editing. This process yields editing samples with changes in style, object, color, material, or expression. Then editing samples generated by these automated pipelines are further filtered with various rules, ultimately producing a total of 3.5 million data.

For the second part of SEED-Data-Edit, we crawl image editing pairs from multiple websites where amateur photographers post their images along with editing requests. These requests are then addressed by Photoshop experts who provide the edited images. To ensure consistency between the instructions and the before-and-after editing images, all editing pairs are manually re-annotated with instructions by human annotators. This process results in a collection of 52K image editing pairs that reflect real-world scenarios. For the third part of SEED-Data-Edit, we employ Photoshop experts to perform a series of multi-turn edits on real images and obtain the editing instructions for each round. This process generates 95K multi-turn editing data, with up to five rounds per editing sequence.

In order to demonstrate the effectiveness of SEED-Data-Edit, we fine-tune a pre-trained Multimodal Large Language Model (MLLM) SEED-X~\cite{ge2024seed} with this dataset and yield the instruction-tuned model SEED-X-Edit. SEED-X unifies comprehension and generation through decoding images from the predicted ViT~\cite{dosovitskiy2020image} features with a pre-trained visual de-tokenizer, which enables it to understand multimodal input (\textit{e.g.}, a source image and a language instruction) and generate a target image after editing. SEED-X-Edit model achieves promising results in language-guided image editing, which showcases the potential of our dataset in advancing this field. It's worth noting that since SEED-X pre-trains the visual de-tokenizer also with SEED-Data-Edit, incorporating the conditional image (\textit{i.e.} a source image) as an additional input besides the high-level image features, its visual de-tokenizer can recover the fine-grained details of the original image (\textit{i.e.} a target image). This characteristic is particularly beneficial for high-precision image editing, where the details of the source image should be preserved.

Through combining large-scale automated pipeline-generated edits, real-world scenario editing examples, and human-annotated multi-turn editing data, we aim to make SEED-Data-Edit a comprehensive and versatile resource for training effective language-guided image editing models. All data of SEED-Data-Edit and the instruction-tuned model SEED-X-Edit are released.

\begin{table}
\small
\centering
\caption{Comparison of existing image editing datasets. ``Real Image for Edit '' denotes whether real images are used for editing instead of images generated by models. ``Real-world Scenario'' indicates whether images edited by users in the real world are included. ``Human'' denotes whether human annotators are involved.}
\label{tab:comparison}
\resizebox{1.0\columnwidth}{!}{
\begin{tabular}{c|cccrcr}
\toprule
                & Real Image for Edit & Real-world Scenario &Human & \# Edits & \# Max Rounds & \# Multi-turn \\
\toprule
InstructPix2Pix &$\times$                        &$\times$         &$\times$              & 313,010  & 1             & 0             \\
MagicBrush      &$\checkmark$                         &$\times$     &$\checkmark$                   & 10,388   & 3             & 3,088         \\
HQ-Edit         &$\times$                        &$\times$          &$\times$              & 197,350  & 1             & 0             \\
SEED-Data-Edit       &$\checkmark$                         & $\checkmark$       &$\checkmark$               & 3,669,644 & 5             & 21,382   \\     
\toprule
\end{tabular}}
\end{table}

\section{Related Work}
Table.~\ref{tab:comparison} compares the existing language-guided image editing datasets including InstructPix2Pix~\cite{brooks2023instructpix2pix}, MagicBrush~\cite{zhang2024magicbrush} and HQ-Edit~\cite{hui2024hq}. InstructPix2Pix adopts Prompt-to-Prompt~\cite{hertz2022prompt} to generate a source image and a target image based on an input caption of LAION-Aesthetics~\cite{laion_aesthetics} dataset and a target caption after editing. However, it only includes single-turn image editing pairs, and all images are model-generated without the inclusion of real images. MagicBrush hires crowd workers on Amazon Mechanical Turk (AMT) to manually annotate images from MS COCO dataset~\cite{lin2014microsoft} using the DALL-E 2 platform~\cite{dall2} for multi-turn editing pairs. However, it only includes 10K pairs with a maximum of three rounds per editing sequence. HQ-Edit first generates image descriptions and edit instructions using GPT-4~\cite{openai2023gpt4}, and then generates diptychs using GPT-4V~\cite{2023GPT4VisionSC} and DALL-E 3~\cite{BetkerImprovingIG} based on image descriptions and instructions. The diptychs are further split into a source image and a target image, with the instruction being rewritten by GPT-4V.  However, the generation of diptychs does not guarantee that the fine-grained details of the source image are preserved in the target image, and lack realism in the generated images. In SEED-Data-Edit, we combine three distinct types of data: large-scale automated pipeline-generated edits, real-world scenario data, and multi-turn editing data annotated by Photoshop experts. The dataset comprises a total of 3.7M image editing pairs and 21K multi-turn editing sequences, with a maximum of 5 rounds per sequence.

\section{Dataset}
As demonstrated in Fig.~\ref{fig:teaser_example} and Fig.~\ref{fig:example}, SEED-Data-Edit is composed of three distinct types of data. (a) Automated pipeline-generated data, which ensures a substantial volume of diverse image editing examples. (b) Real-world scenario data collected from websites where amateur photographers post their images along with editing requests, which are then addressed by Photoshop experts. This part of data captures the intricacies of user intentions and promotes the practical application of image editing in real-life situations. (c) Multi-turn editing data, which involves multiple rounds of edits performed by Photoshop experts on real images, simulating iterative editing processes.

\begin{figure}[h!]
\centering
\makebox[\textwidth][c]{\includegraphics[width=1.0\linewidth]{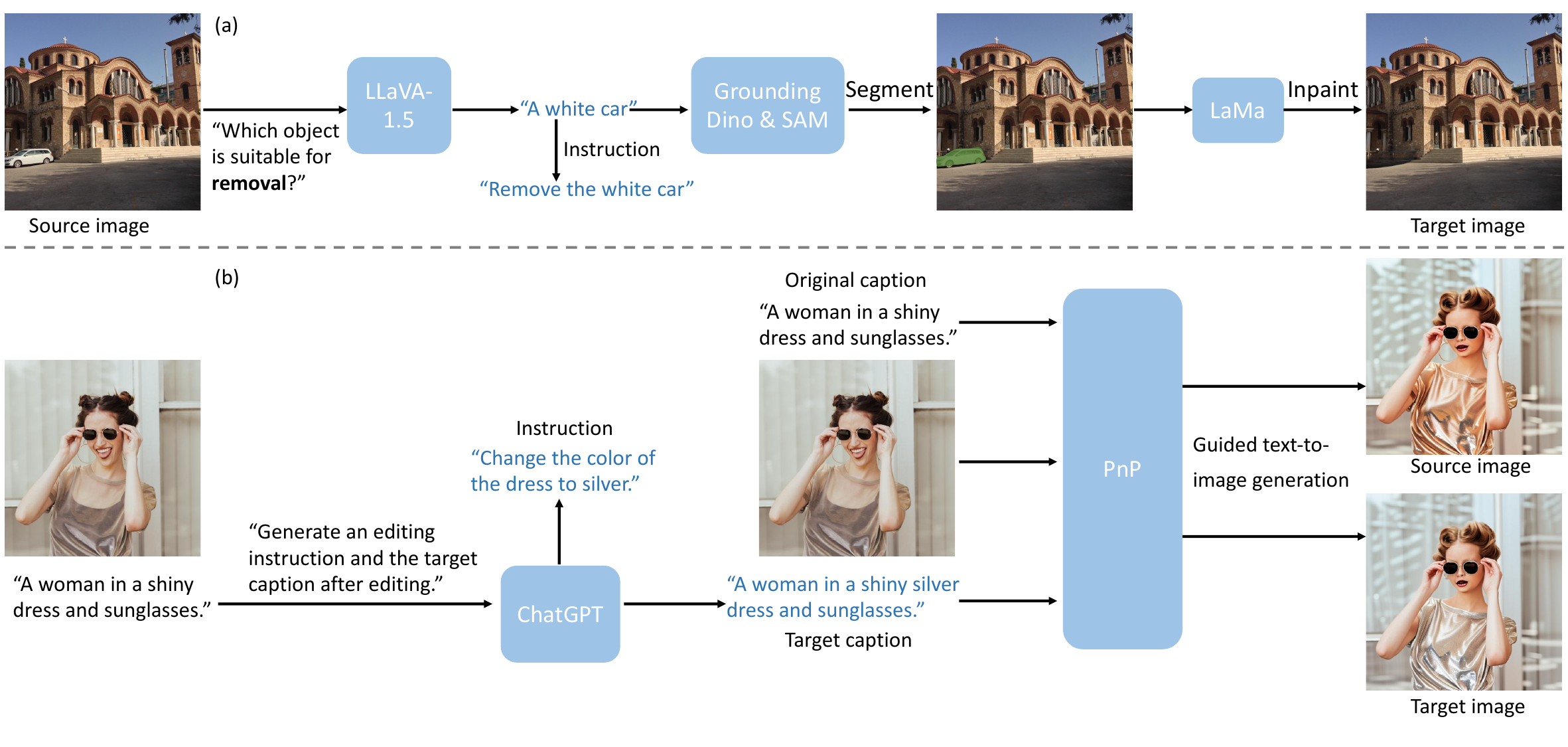}}%
\vspace{-5pt}
\caption{The automated pipelines for generating editing pairs, which constitutes the first part of SEED-Data-Edit. In pipeline (a), an object is first segmented and subsequently removed with an inpainting module, which results in a set of ``Remove'' and ``Add'' editing samples.  In pipeline (b), an image-guided text-to-image generation model is utilized to generate source images and target images based on the original image, the original caption and the target caption after editing.}
\label{fig:pipeline}
\vspace{-10pt}
\end{figure}

\subsection{Automated Pipeline-generated Data}
We adopt two automated pipelines to generate a large number of high-quality image editing pairs as shown in Fig.~\ref{fig:pipeline}. Pipeline (a) is designed to create a set of ``Remova'' edits, where ``Add'' samples are generated by reversing the ``Remove'' operation, as it is difficult to directly add an object in a suitable location in the image. Pipeline (b) is designed to produce editing samples with changes in style, object, color, material, or expression. 

Specifically, in pipeline (a) given a real image, LLaVA-1.5~\cite{liu2023improved} is employed to answer the question ``Which object is suitable for removal?''. Then GroundingDino~\cite{liu2023grounding} and SAM~\cite{kirillov2023segany} are employed to segment the target object with the mask. Given the original image and the segmentation mask, an inpainting model LaMa~\cite{suvorov2021resolution} is utilized to inpaint the image, removing the target object and creating the target image. The corresponding instruction is produced through filling in the template ``Remove something from the image'' with the target object. To generate the ``Add'' samples, we reverse the process by using the target image as the source image. In this case, the instruction becomes ``Add something in the image'' and the source image serves as the actual target image. This approach ensures that the added object is placed in an appropriate location within the image, maintaining visual coherence and realism in the editing process. We filter out data where the size of the target object exceeds a certain threshold. We use images from Unsplash~\cite{unsplash} for generating image editing pairs with pipeline (a), and produce a total of 1.5M image editing pairs after filtering.

In pipeline(b), given a caption of an input image, we first employ ChatGPT~\cite{ChatGPT} to generate an editing instruction and the target instruction after the editing has been executed. Subsequently, we utilize an
image-guided text-to-image generation model PnP~\cite{Tumanyan_2023_CVPR} to generate a target image given the original image and target caption. To ensure the visual consistency between the target image and image before editing, we further utilize PnP to generate the source image given the original image and original caption. We filter the data by calculating the CLIP~\cite{radford2021learning} similarity between the target image and the target caption, as well as between the target image and the original caption. If the latter similarity is higher than the former, the data will be filtered out, since the visual content of the target image does not accurately reflect the editing instruction. We use images from Openimages~\cite{kuznetsova2020open} for generating image editing pairs with pipeline (b), and produce a total of 2.0M image editing pairs after filtering.

\subsection{Real-world Scenario Data}
\begin{figure}[h!]
\vspace{5pt}
\centering
\makebox[\textwidth][c]{\includegraphics[width=1.1\linewidth]{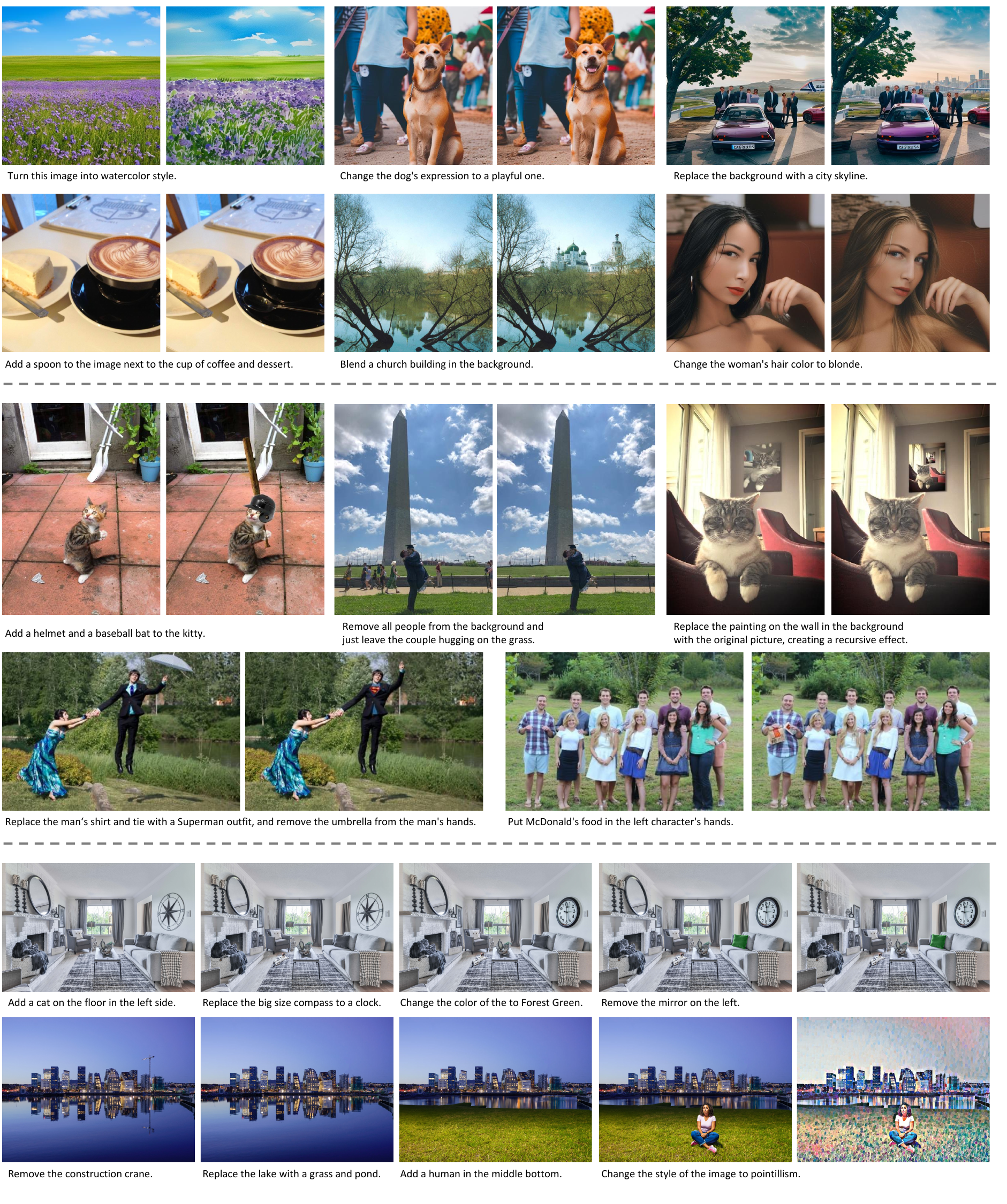}}%
\caption{More examples of instruction-guided image editing in SEED-Data-Edit, which integrates three distinct types of data: (a) Large-scale automated pipeline-generated edits (first and second row), (b) Real-world scenario data, where
amateur photographers post their images along with editing requests, which are addressed by Photoshop experts (third and fourth row), and (c) Multi-turn editing data annotated by Photoshop experts on real images (fifth and sixth row).}
\label{fig:example}
\end{figure}

We crawl image editing pairs from four websites~\cite{photoshopbattles, photoshopgurus, PhotoshopRequest, zhopped}, where amateur photographers post their images accompanied by editing requests. These requests are then fulfilled by Photoshop experts who provide the edited images as target images. To ensure that the editing instruction can accurately reflect the change from the source image to the target image, we employ human annotators to re-annotate all editing pairs. This process yields a total of 52K image editing samples. As depicted in Fig.~\ref{fig:teaser_example} and Fig.~\ref{fig:example}, this part of data is more complex, and the instructions are more imaginative, effectively reflecting real-world user image editing intentions. 

\subsection{Multi-turn Editing Data}
We employ Photoshop experts to edit images in sequence using Photoshop for multi-turn image edits, documenting the editing instruction for each round. Each editing round encompasses various modifications, including (1) replacing an object, (2) adding an object, (3) removing an object, (4) changing the action, (5) altering text or patterns, and (6) modifying the count of objects. To ensure the diversity of the input image, we use images from Unsplash~\cite{unsplash}, SAM~\cite{kirillov2023segany} and JourneyDB~\cite{sun2024journeydb} for multi-turn editing. In total, we generate 21K multi-turn rounds with a maximum of 5 rounds, resulting in 95K image editing pairs.

\begin{figure}[h!]
\centering
\makebox[\textwidth][c]{\includegraphics[width=1.05\linewidth]{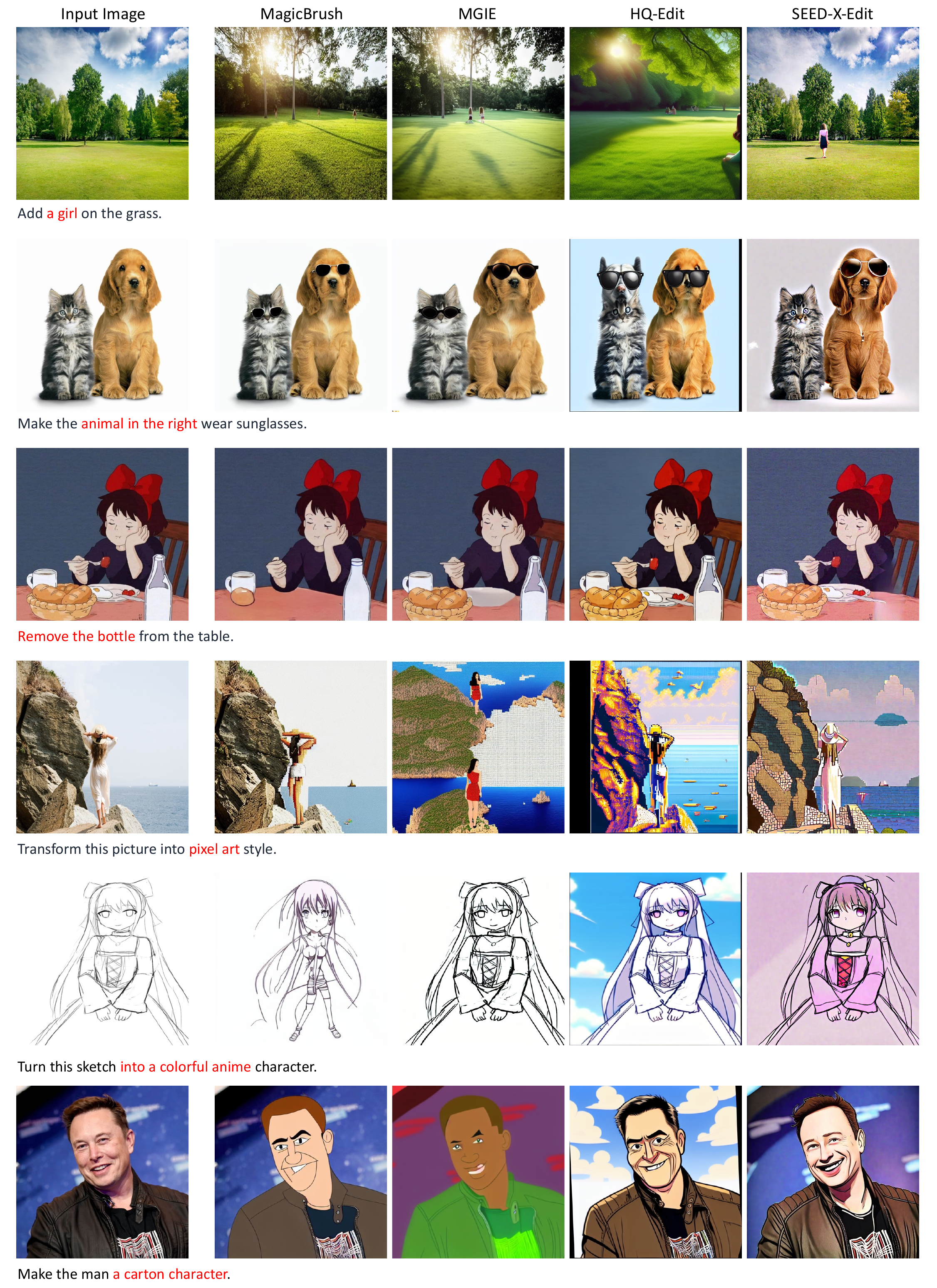}}%
\caption{The comparison of language-guided image editing between existing methods and SEED-X-Edit. Through find-tuning with SEED-Data-Edit, SEED-X-Edit is able to adhere to editing instructions more accurately.}
\label{fig:comparison}
\end{figure}

\section{SEED-X-Edit}
We fine-tune a pre-trained Multimodal Large Language Model (MLLM) SEED-X (\url{https://github.com/AILab-CVC/SEED-X})~\cite{ge2024seed} with SEED-Data-Edit, InstructPix2Pix~\cite{brooks2023instructpix2pix}, and MagicBrush~\cite{zhang2024magicbrush}, which yields the instruction-tuned model SEED-X-Edit. SEED-X unifies multimodal comprehension and generation through decoding images from the predicted ViT~\cite{dosovitskiy2020image} features with a pre-trained visual de-tokenizer. In this way, it is able to comprehend multimodal input (\textit{e.g.}, a source image and a language editing), and generate a target image after decoding. We fine-tune SEED-X using a LoRA~\cite{hu2021lora} module with 16 A100-40G GPUs, which takes around 40 hours. Note that we do not fine-tune with multi-turn editing sequences and instead use multi-turn editing data in a single-turn way, and we will explore multi-turn image editing in future work.

We compare SEED-X-Edit with the baseline models including MagicBrush~\cite{zhang2024magicbrush}, MGIE~\cite{fu2023guiding}, and HQ-Edit~\cite{hui2024hq}. Here, MGIE model is trained on InstructPix2Pix data, MagicBrush model is fine-tuned from an InstructPix2Pix trained model with MagicBrush data. As illustrated in Fig.~\ref{fig:comparison}, SEED-X-Edit model achieves promising results in language-guided image editing and can better adhere to editing instructions compared to models trained with existing image editing datasets. For example, SEED-X-Edit is able to add sunglasses to the animal in the right while all baseline models put sunglasses on both animals. Additionally, SEED-X-Edit successfully removes the bottle from the table while the baseline models fail to execute this editing instruction. The satisfactory results showcases the potential of our dataset in advancing instructional image editing. It is worth noting that SEED-X also pre-trains the visual de-tokenizer with SEED-Data-Edit, which incorporates the source image as an additional input besides the ViT features of the target images to reconstruct the target image. This characteristic is particularly beneficial for high-precision image editing, as it enables the preservation of the source image's fine-grained details after editing.

\section{Conclusion}
In this technical report, we introduce SEED-Data-Edit, which is a unique hybrid dataset for language-guided image editing. SEED-Data-Edit integrates three distinct types of data, including large-scale automated pipeline-generated edits, real-world scenario editing examples, and human-annotated multi-turn editing data. We further train a language-guided image editing model named SEED-X-Edit with our dataset from a pre-trained MLLM SEED-X, and SEED-X-Edit achieves promising results. All data of SEED-Data-Edit and the instruction-tuned model SEED-X-Edit are released.

{\small
\bibliographystyle{unsrt}
\bibliography{SEED-Edit}

\begin{thebibliography}{10}

\bibitem{zhang2024magicbrush}
Kai Zhang, Lingbo Mo, Wenhu Chen, Huan Sun, and Yu~Su.
\newblock Magicbrush: A manually annotated dataset for instruction-guided image editing.
\newblock {\em Advances in Neural Information Processing Systems}, 36, 2024.

\bibitem{brooks2023instructpix2pix}
Tim Brooks, Aleksander Holynski, and Alexei~A Efros.
\newblock Instructpix2pix: Learning to follow image editing instructions.
\newblock In {\em Proceedings of the IEEE/CVF Conference on Computer Vision and Pattern Recognition}, pages 18392--18402, 2023.

\bibitem{hui2024hq}
Mude Hui, Siwei Yang, Bingchen Zhao, Yichun Shi, Heng Wang, Peng Wang, Yuyin Zhou, and Cihang Xie.
\newblock Hq-edit: A high-quality dataset for instruction-based image editing.
\newblock {\em arXiv preprint arXiv:2404.09990}, 2024.

\bibitem{sheynin2023emu}
Shelly Sheynin, Adam Polyak, Uriel Singer, Yuval Kirstain, Amit Zohar, Oron Ashual, Devi Parikh, and Yaniv Taigman.
\newblock Emu edit: Precise image editing via recognition and generation tasks.
\newblock {\em arXiv preprint arXiv:2311.10089}, 2023.

\bibitem{fu2023guiding}
Tsu-Jui Fu, Wenze Hu, Xianzhi Du, William~Yang Wang, Yinfei Yang, and Zhe Gan.
\newblock Guiding instruction-based image editing via multimodal large language models.
\newblock {\em arXiv preprint arXiv:2309.17102}, 2023.

\bibitem{rombach2022high}
Robin Rombach, Andreas Blattmann, Dominik Lorenz, Patrick Esser, and Bj{\"o}rn Ommer.
\newblock High-resolution image synthesis with latent diffusion models.
\newblock In {\em Proceedings of the IEEE/CVF conference on computer vision and pattern recognition}, pages 10684--10695, 2022.

\bibitem{betker2023improving}
James Betker, Gabriel Goh, Li~Jing, Tim Brooks, Jianfeng Wang, Linjie Li, Long Ouyang, Juntang Zhuang, Joyce Lee, Yufei Guo, et~al.
\newblock Improving image generation with better captions.
\newblock {\em Computer Science. https://cdn. openai. com/papers/dall-e-3. pdf}, 2(3):8, 2023.

\bibitem{chen2023pixart}
Junsong Chen, Jincheng Yu, Chongjian Ge, Lewei Yao, Enze Xie, Yue Wu, Zhongdao Wang, James Kwok, Ping Luo, Huchuan Lu, et~al.
\newblock Pixart: Fast training of diffusion transformer for photorealistic text-to-image synthesis.
\newblock {\em arXiv preprint arXiv:2310.00426}, 2023.

\bibitem{ge2024seed}
Yuying Ge, Sijie Zhao, Jinguo Zhu, Yixiao Ge, Kun Yi, Lin Song, Chen Li, Xiaohan Ding, and Ying Shan.
\newblock Seed-x: Multimodal models with unified multi-granularity comprehension and generation.
\newblock {\em arXiv preprint arXiv:2404.14396}, 2024.

\bibitem{dosovitskiy2020image}
Alexey Dosovitskiy, Lucas Beyer, Alexander Kolesnikov, Dirk Weissenborn, Xiaohua Zhai, Thomas Unterthiner, Mostafa Dehghani, Matthias Minderer, Georg Heigold, Sylvain Gelly, et~al.
\newblock An image is worth 16x16 words: Transformers for image recognition at scale.
\newblock {\em arXiv preprint arXiv:2010.11929}, 2020.

\bibitem{hertz2022prompt}
Amir Hertz, Ron Mokady, Jay Tenenbaum, Kfir Aberman, Yael Pritch, and Daniel Cohen-Or.
\newblock Prompt-to-prompt image editing with cross attention control.
\newblock {\em arXiv preprint arXiv:2208.01626}, 2022.

\bibitem{laion_aesthetics}
Christoph Schuhmann and Romain Beaumont.
\newblock Laion-aesthetics.
\newblock \url{https://laion.ai/blog/laion-aesthetics/}, 2022.

\bibitem{lin2014microsoft}
Tsung-Yi Lin, Michael Maire, Serge Belongie, James Hays, Pietro Perona, Deva Ramanan, Piotr Doll{\'a}r, and C~Lawrence Zitnick.
\newblock Microsoft coco: Common objects in context.
\newblock In {\em Computer Vision--ECCV 2014: 13th European Conference, Zurich, Switzerland, September 6-12, 2014, Proceedings, Part V 13}, pages 740--755. Springer, 2014.

\bibitem{dall2}
Dall-e 2.
\newblock \url{https://openai.com/dall-e-2}, 2022.

\bibitem{openai2023gpt4}
OpenAI.
\newblock Gpt-4 technical report, 2023.

\bibitem{2023GPT4VisionSC}
Gpt-4v(ision) system card.
\newblock 2023.

\bibitem{BetkerImprovingIG}
James Betker, Gabriel Goh, Li~Jing, TimBrooks, Jianfeng Wang, Linjie Li, LongOuyang, JuntangZhuang, JoyceLee, YufeiGuo, WesamManassra, PrafullaDhariwal, CaseyChu, YunxinJiao, and Aditya Ramesh.
\newblock Improving image generation with better captions.

\bibitem{liu2023improved}
Haotian Liu, Chunyuan Li, Yuheng Li, and Yong~Jae Lee.
\newblock Improved baselines with visual instruction tuning.
\newblock {\em arXiv preprint arXiv:2310.03744}, 2023.

\bibitem{liu2023grounding}
Shilong Liu, Zhaoyang Zeng, Tianhe Ren, Feng Li, Hao Zhang, Jie Yang, Chunyuan Li, Jianwei Yang, Hang Su, Jun Zhu, et~al.
\newblock Grounding dino: Marrying dino with grounded pre-training for open-set object detection.
\newblock {\em arXiv preprint arXiv:2303.05499}, 2023.

\bibitem{kirillov2023segany}
Alexander Kirillov, Eric Mintun, Nikhila Ravi, Hanzi Mao, Chloe Rolland, Laura Gustafson, Tete Xiao, Spencer Whitehead, Alexander~C. Berg, Wan-Yen Lo, Piotr Doll{\'a}r, and Ross Girshick.
\newblock Segment anything.
\newblock {\em arXiv:2304.02643}, 2023.

\bibitem{suvorov2021resolution}
Roman Suvorov, Elizaveta Logacheva, Anton Mashikhin, Anastasia Remizova, Arsenii Ashukha, Aleksei Silvestrov, Naejin Kong, Harshith Goka, Kiwoong Park, and Victor Lempitsky.
\newblock Resolution-robust large mask inpainting with fourier convolutions.
\newblock {\em arXiv preprint arXiv:2109.07161}, 2021.

\bibitem{unsplash}
Zahid Ali, Chesser Luke, and Carbone Timothy.
\newblock Unsplash.
\newblock \url{https://github.com/unsplash/datasets}, 2023.

\bibitem{ChatGPT}
OpenAI.
\newblock Introducing chatgpt.
\newblock https://openai.com/blog/chatgpt, 2022.

\bibitem{Tumanyan_2023_CVPR}
Narek Tumanyan, Michal Geyer, Shai Bagon, and Tali Dekel.
\newblock Plug-and-play diffusion features for text-driven image-to-image translation.
\newblock In {\em Proceedings of the IEEE/CVF Conference on Computer Vision and Pattern Recognition (CVPR)}, pages 1921--1930, June 2023.

\bibitem{radford2021learning}
Alec Radford, Jong~Wook Kim, Chris Hallacy, Aditya Ramesh, Gabriel Goh, Sandhini Agarwal, Girish Sastry, Amanda Askell, Pamela Mishkin, Jack Clark, et~al.
\newblock Learning transferable visual models from natural language supervision.
\newblock In {\em International conference on machine learning}, pages 8748--8763. PMLR, 2021.

\bibitem{kuznetsova2020open}
Alina Kuznetsova, Hassan Rom, Neil Alldrin, Jasper Uijlings, Ivan Krasin, Jordi Pont-Tuset, Shahab Kamali, Stefan Popov, Matteo Malloci, Alexander Kolesnikov, et~al.
\newblock The open images dataset v4: Unified image classification, object detection, and visual relationship detection at scale.
\newblock {\em International journal of computer vision}, 128(7):1956--1981, 2020.

\bibitem{photoshopbattles}
Photoshopbattles.
\newblock https://www.reddit.com/r/photoshopbattles/, 2024.

\bibitem{photoshopgurus}
Photoshop gurus.
\newblock https://www.photoshopgurus.com/forum/, 2024.

\bibitem{PhotoshopRequest}
Photoshoprequest.
\newblock https://www.reddit.com/r/PhotoshopRequest/, 2024.

\bibitem{zhopped}
Zhopped.
\newblock http://zhopped.com/, 2024.

\bibitem{sun2024journeydb}
Keqiang Sun, Junting Pan, Yuying Ge, Hao Li, Haodong Duan, Xiaoshi Wu, Renrui Zhang, Aojun Zhou, Zipeng Qin, Yi~Wang, et~al.
\newblock Journeydb: A benchmark for generative image understanding.
\newblock {\em Advances in Neural Information Processing Systems}, 36, 2024.

\bibitem{hu2021lora}
Edward~J Hu, Yelong Shen, Phillip Wallis, Zeyuan Allen-Zhu, Yuanzhi Li, Shean Wang, Lu~Wang, and Weizhu Chen.
\newblock Lora: Low-rank adaptation of large language models.
\newblock {\em arXiv preprint arXiv:2106.09685}, 2021.

\end{thebibliography}
}

\end{document}